# Task tree retrieval from FOON using search algorithms

Amitha Attapu
*Department of Computer Science and Engineering*
*University of South Florida*
Tampa FL, USA
amithaattapu@usf.edu

*Abstract*—Robots use Functional object-oriented network(FOON) to carry out a given task by following the steps in it. It is a knowledge representation network. We are given a Universal FOON, which contains many cooking recipes. Given a goal node (goal output recipe), we retrieve a task tree from universal FOON, which gives the steps to be followed to reach the goal node. We do this using two search algorithms, IDS (Iterative deepening search) and GBFS(Greedy best-first search). We implement the GBFS with two different heuristic functions. We compare which algorithm is giving the best results. We observe that for the goal nodes considered for the experiment, different algorithms are giving the best performance in different cases.

## I. Introduction

Robots can be very useful to automate tasks and reduce the human effort required. But for the robot to know, how to perform tasks, we need to give it a clear set of steps to follow. It is nearly impossible to provide a robot with instructions for every possible task. Therefore we have a Universal Functional object-oriented network (FOON) which was created and expanded and has a lot of existing recipe information [1].

But certain tasks are complicated for robots to perform and similarly, some tasks are complicated for humans to perform. Therefore weights have been added to functional units to represent the chance of successful execution of the motion by the robot [2].

Given a set of kitchen items and a goal node, using Universal FOON, a robot must be able to determine if the required items are present in the kitchen, and if yes, get the steps to convert the required kitchen items to the goal node. Now through this paper, we use two algorithms (IDS and GBFS) to retrieve a task tree (if possible) for a goal node and a given set of kitchen items.

The following would be the different parts of the paper: Section II FOON creation, where we will discuss the different terminologies related to FOON and visualization of FOON. In Section III Methodology we discuss the IDS and GBFS search algorithms and the two different heuristics implemented and used in GBFS. In Section IV Experiment/Discussion, we compare the performance of different algorithms. In the final section V, we specify the references of the papers that have been cited.

## II. FOON CREATION

### A. Terminologies

FOON: It is a bipartite network. It has input nodes which get converted to output nodes when they go through a motion node. The motion node will result in the change of state of input nodes to output nodes [3].

Node: There are three kinds of nodes, input nodes, output nodes, and motion nodes [3].

Edges: The FOON is a directed graph, and the edges represent whether the given nodes are input nodes or output nodes. The nodes that are coming into the motion node are input nodes, whereas the ones that are going out from motion nodes are output nodes [3].

Functional unit: The relationship between one or more items and a single functional motion related to those objects is represented by this [3].

Task tree: It is a tree generated by the algorithms for a given goal node and a given set of kitchen items. If the goal node(final result) can be reached using the ingredients in the kitchen, then the task tree is generated.

Universal FOON: It is a graph, which has been formed after merging multiple recipe sub-graphs [3].

*FIG .1. ICE task tree*

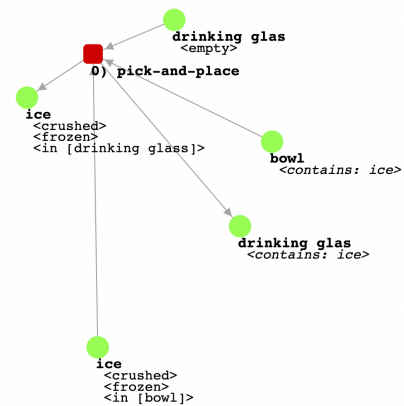

*B. Task tree creation process*

For the task tree creation, we preprocessed the data from FOON.txt in preprocess.py, and created a graph for universal FOON. We used the kitchen.json file to get the list of items in the kitchen. We load the goal nodes from the goal_nodes.json file provided. We have a FOON class file that specifies the attributes of FOON, such as states, and ingredients.

The main file for the search algorithms is search.py. We have implemented all three search algorithms here. From the main method of search.py, we load goal nodes and call each of the three algorithms for each goal node, and save the resultant task tree into a .txt file.

We have discussed the algorithms implementation in the next section.

To check if there are any mistakes or disconnectivity in the task tree, we used the visualizer in the foonets website. We uploaded the .txt file of the task tree retrieved to create the visualization.

*FIG .2. Universal FOON visualization*

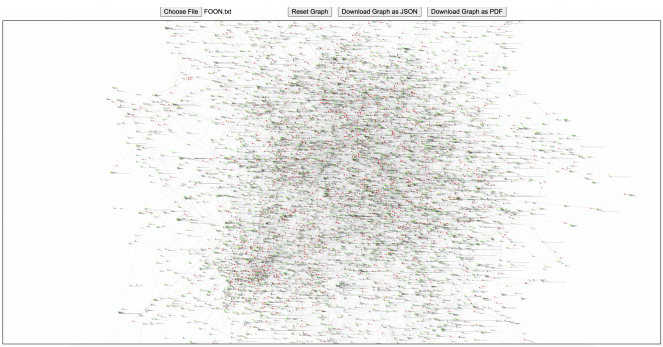

### III. METHODOLOGY

We have implemented 3 different algorithms, IDS, GBFS (with two different heuristics) for task tree retrieval. We discuss below the implementation of the algorithms.

*A. Iterative Deepening Search (IDS)*

IDS search is similar to depth-first search, with a bound on depth. On each each iteration we increase the depth until we find the desired answer.

Coming to our implementation, we start the search from the goal node and work our way upwards. I.e, We first search for the functional unit which has the goal node first, and on finding it we search for the input nodes of the functional unit. If all of them are already found in the kitchen, our search ends there. Otherwise, we further search each input node. Once all the input nodes are found, we reverse the task tree and return the resultant task tree.

In IDS, we do a depth-first search, with a depth bound each time. We start from an initial depth of 0 and increase our depth until we find the complete task tree.

For implementing IDS, we have used recursion. We have implemented a recursive method IDS, which takes in 4 inputs: task tree, goal id being searched, kitchen items, and current depth. This method returns a boolean. If task tree is found, it returns true else false.

We call this recursive method from an external function, with depth starting at 0. Until the function returns True, we keep increasing the depth and call it again.

Inside the IDS, we have the following conditions: we check if the depth has gone below 1, and if yes we return False. Else, we check if the item already exists in the kitchen, if yes we return True. If not, we search for functional units which have the output node as the one we are searching. Now we start iterating through the input nodes of this functional unit, and recursively call IDS to search for the nodes. Note that, we decrease the depth by 1, every time we make the recursive call.

*B. Greedy best first search*

- It is an informed search algorithm, and if we have multiple nodes to choose the next move from, we use a heuristic function to determine the next move.

- In our implementation, we have a list which contains items_to_search. Initially, we only have the goal node in it. Now we have a while loop, which iterates until the size of items_to_search is greater than 0. We check if the goal node is in the kitchen, if yes, our search ends there. Otherwise, we find functional units which output the goal node. If we find multiple functional units, we use the heuristic function to decide which function unit to choose. Once chosen we iterate through its inputs, and add them to items to the search list, and our process continues until items_to_search list is empty

*FIG .3. Functional unit*

```
1   //
2   O   drinking glas
3   S   empty
4   O   bucket
5   S   contains    {ice}
6   O   ice
7   S   crushed
8   S   frozen
9   S   in  [bowl]
10  O   measuring cup
11  S   empty
12  M   scoop and pour
13  O   drinking glas
14  S   contains    {ice}
15  O   ice
16  S   crushed
17  S   frozen
18  S   in  [drinking glass]
19  //
```

*C. Heuristic Functions*

Heuristic a: We choose the functional unit, whose motion node has the highest success rate.

Heuristic b: We chose the functional unit which has the least number of input nodes.

## IV. EXPERIMENT/DISCUSSION

We have tested the three algorithms (IDS, GBFS heuristic 1, and GBFS heuristic 2) using 6 goal nodes. We have searched for each of the goal nodes, from all three algorithms, Table. 1 shows the number of functional units obtained during each algorithm.

I. NUMBER OF FUNCTIONAL UNITS FOR DIFFERENT GOAL NODES VS SEARCH ALGORITHMS

| Goal node | No of functional units | | |
|---|---|---|---|
| | IDS | GBFS Heuristic 1 | GBFS Heuristic 2 |
| Greek salad | 26 | 33 | 26 |
| Macaroni | 7 | 7 | 8 |
| Ice | 1 | 1 | 1 |
| Sweet potato | 3 | 3 | 3 |
| Whipped cream | 10 | 10 | 14 |
| Carrot salad | 34 | 31 | 34 |

From the table, we observe that in the case of greek salad, IDS and GBFS with heuristic 2 have performed better than GBFS with heuristic 1, as they have reached the goal node, with a lesser number of functional units. This implies that they have a lesser number of steps in the recipe to reach the goal node.

In the case of Macaroni goal node, IDS, and GBFS with heuristic 1 have performed better than GBFS with heuristic 2, as they have reached the goal node, with a lesser number of functional units. Coming to the case of Ice and sweet potato, all three algorithms have performed equally.

We see some difference in the performance in the case of whipped cream, IDS and GBFS with heuristic 1 have performed better. In the case of carrot salad, GBFS with heuristic 1 has outperformed IDS and GBFS with heuristic 2.

As we can see different algorithms are performing better for different goal nodes. Based on the experiments conducted, no particular algorithm could be concluded to be the better one. But based on completeness, time complexity, and memory complexity we can draw some conclusions.

The memory complexity of IDS is $O(b*d)$, where b is the branching factor and d is the depth. The worst-case memory complexity of GBFS is $O(b^m)$, where 'm' is the maximum depth. But the performance of GBFS hugely depends on the heuristic function and a good heuristic function can improve the performance. The IDS algorithm is complete, i.e. it will always find the optimal solution when given enough depth. Whereas the GBFS might choose the wrong node to explore and might get stuck at a local minimum.

The time complexity of IDS is $O(b^d)$ and the worst-case time complexity for GBFS is $O(b^m)$, but a good heuristic function affects and improves the performance a lot.